\documentclass{article} 
\usepackage{iclr2025_conference,times}

\usepackage{amsmath,amsfonts,bm}

\def\eqref#1{equation~\ref{#1}}

\def\1{\bm{1}}

\DeclareMathAlphabet{\mathsfit}{\encodingdefault}{\sfdefault}{m}{sl}
\SetMathAlphabet{\mathsfit}{bold}{\encodingdefault}{\sfdefault}{bx}{n}

\usepackage{hyperref}
\usepackage{url}
\usepackage{algorithm}
\usepackage{algorithmic}

\usepackage{commath}
\usepackage{amsmath}
\usepackage{amssymb}
\usepackage{amsthm}
\usepackage{bm}
\usepackage{enumitem}
\usepackage{graphicx}
\usepackage{subcaption}
\usepackage{booktabs}
\usepackage{adjustbox}

\usepackage{mdframed}
\usepackage{tcolorbox}

\title{Boosting Deductive Reasoning \\ with Step Signals In RLHF}

\author{Jialian Li$^{1}$\thanks{Equal contribution.}, Yipin Zhang$^{1}$\footnotemark[1], Wei Shen$^{1}$, Yuzi Yan$^{1, 2}$\thanks{Interns at Baichuan}, Jian Xie$^{1}$, Dong Yan$^{1}$\thanks{Corresponding authors.} \\
        \textsuperscript{1}Baichuan AI \\
        \textsuperscript{2}Department of Electronic Engineering, Tsinghua University \\
        lijialian7@163.com,\;\ zypzyp665@gmail.com,\;\ weyshiocn@gmail.com,\;\\
        yan-yz17@tsinghua.org.cn,\;\ xiejian1990@gmail.com,\;\ sproblvem@gmail.com
        }

\iclrfinalcopy 
\begin{document}

\maketitle

\begin{abstract}

Logical reasoning is a crucial task for Large Language Models (LLMs), enabling them to tackle complex problems. Among reasoning tasks, multi-step reasoning poses a particular challenge. Grounded in the theory of formal logic, we have developed an automated method, Multi-step Deduction (MuseD), for deductive reasoning data. MuseD has allowed us to create training and testing datasets for multi-step reasoning. Our generation method enables control over the complexity of the generated instructions, facilitating training and evaluation of models across different difficulty levels. Through RLHF training, our training data has demonstrated significant improvements in logical capabilities for both in-domain of out-of-domain reasoning tasks. Additionally, we have conducted tests to assess the multi-step reasoning abilities of various models.

\end{abstract}

\section{Introduction}








Recent advancements in large language models (LLMs) \citep{ouyang2022training, 2022Training} have yielded remarkable outcomes. Among the various capabilities of LLMs, reasoning stands out as one of crucial skills, serving as a foundational ability required for solving complex tasks. Numerous efforts~\citep{sun2023survey} have been made to explore and enhance the reasoning capabilities of LLMs.

In this work, we focus on multi-step deductive reasoning tasks~\citep{sun2023survey} within the realm of reasoning. Many previous works \citep{han2024folionaturallanguagereasoning,PrOntoQA,PrOntoQAOOD} have been done of deductive reasoning data generation. However, most works concentrate on supervised fine-tuning \citep{sanh2022multitask} or evaluation. Our work mainly concentrates on generation data for Reinforcement Learning from Human Feedback (RLHF) \citep{ouyang2022training}. Deductive reasoning tasks concentrate on deriving correct conclusions from given premises through rigorous and effective reasoning. Our attention is directed towards constructing high-quality data to improve the deductive reasoning abilities of models during the alignment phase. The inherent rigor of deductive reasoning tasks dictates that the corresponding prompts should not contain contradictory information or be incapable of leading to the correct answer. This presents our first challenge: how to obtain prompts with correct answers and no contradictions. Based on the proper conditions in prompts, we expect LLMs to employ multi-step reasoning to deduce the correct answer. Assessing the correctness of such a multi-step reasoning process constitutes our second issue, since we need accurate scores of responses to construct training data. Lastly, how to efficiently acquire a substantial amount of data for training is the third problem we need to consider.

\begin{figure}[hbt]
    \centering
    \includegraphics[width=0.7\textwidth]{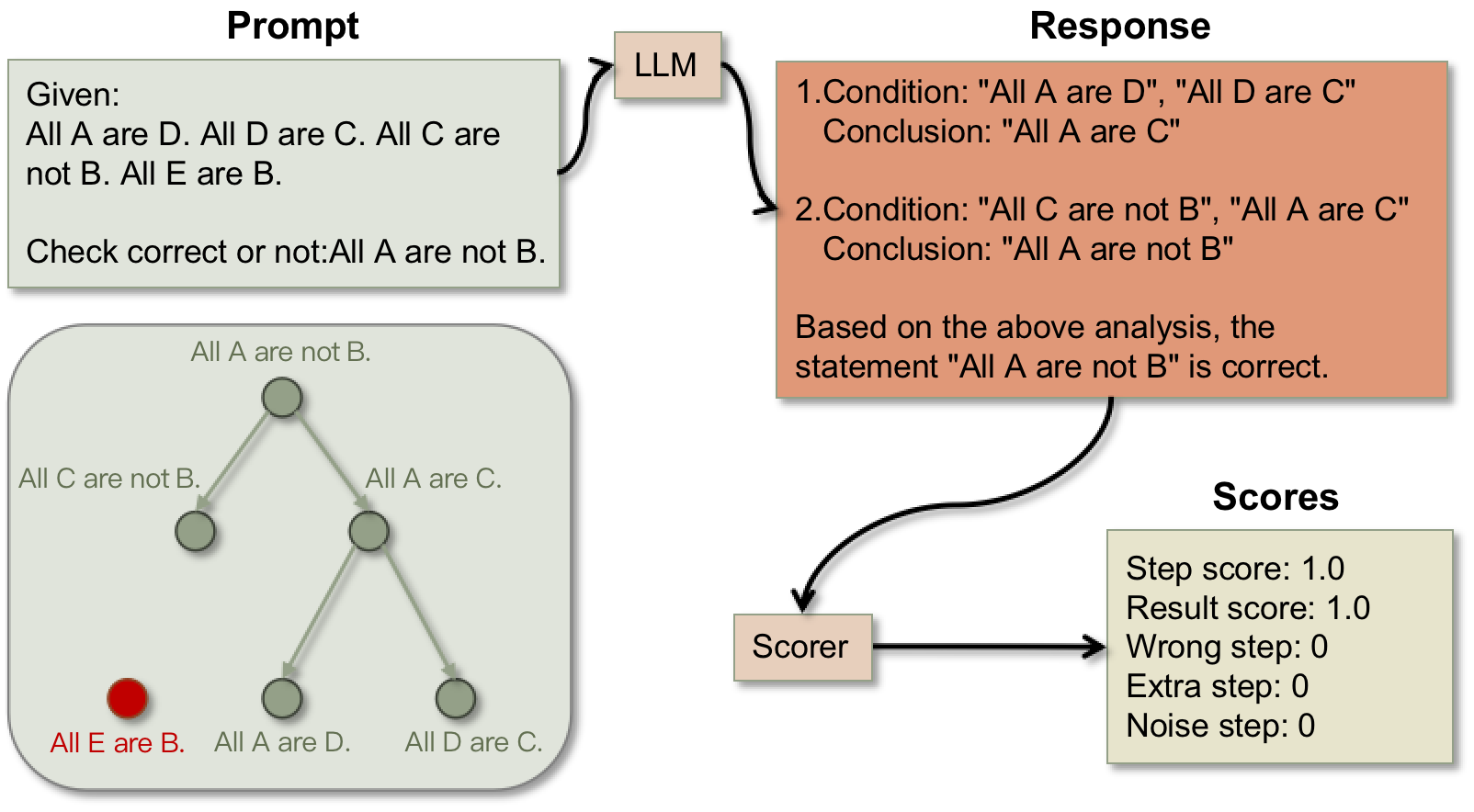}
    \caption{The logic training data pipline.}
    \label{fig:data_pipline}
\end{figure}

To address these issues, we propose a generation scheme for multi-step deductive reasoning data, named Multi-step Deduction (MuseD). MuseD is a scalable approach from prompt creation to final evaluation. To sum up, we base our method MuseD on the syllogistic reasoning of deductive inference~\citep{copi2016introduction}, employing a backward generation approach to obtain the conditions required for the prompt. This ensures that the conditions in the prompt can lead to the correct conclusion without any contradictory conditions. Moreover, by controlling the number of generated conditions, we regulate the number of inference steps required for the prompt. Based on prompts generated by this method, we score the responses of LLMs step by step. That is, we provide an evaluation method that can assess whether an answer was correctly obtained through multi-step reasoning rather than merely being the correct answer. The data generation process is shown in Fig.~\ref{fig:data_pipline}.

Using MuseD, we synthesize partially ordered data for multi-step deductive reasoning and use this data for Reinforcement Learning from Human Feedback (RLHF) training. We achieve significant performance enhancements on both in-domain and out-of-domain reasoning datasets, validating the effectiveness of our synthesized data. Further experiments demonstrate that step scoring is crucial for improving the model's performance in RLHF and that positive rewards for correct answers are the primary motivator for model learning.

Additionally, we utilize our method to create a multi-step deductive reasoning evaluation set, also named MuseD\footnote{The code for MuseD evaluation is given in: https://github.com/zhangyipin/mused/tree/main.}. Compared to previous evaluation sets, it can provide insights into the model's performance changes under tasks with different numbers of reasoning steps and offer more granular process evaluations. In summary, our main contributions include: (1) proposing a data synthesis method MuseD based on multi-step deductive reasoning, (2) achieving improvements in deductive reasoning performance on RLHF with this method, and (3) presenting a multi-step deductive reasoning evaluation set that allows for multi-dimensional automatic scoring.

We will briefly introduce the logical concepts involved in our method in Sec.~\ref{sec:preliminary} and describe our data generation approach in Sec.~\ref{sec:mused_method}. In Sec.~\ref{sec:experiments}, we will present the experimental results, and in Sec.~\ref{sec:evaluation_mused}, we will show the performances of some LLMs on our evaluation set.

\section{Related Work}
\label{sec:related_work}
Many recent works concentrate on the reasoning ability of LLM. \cite{sun2023survey} give a survey on LLM reasoning, where different kinds of reasoning are considered. Traditionally, reasoning are divided into deductive reasoning~\citep{PrOntoQA}, inductive reasoning~\citep{wang2024hypothesis} and abductive reasoning~\citep{bhagavatulaabductive}.

Our work concentrates on deductive reasoning, where valid reasoning process are focused. Many efforts have been done on this tasks to provide high-quality datasets. \cite{tafjord2021proofwriter} give a pipline to generate a synthetic data based on rules. \cite{liu2020logiqa} extend human logical testing to LLM dataset. \cite{han2024folionaturallanguagereasoning} concentrate on natural logical data that are created by experts. \cite{PrOntoQA} and \cite{PrOntoQAOOD} use deduction rules to generate synthetic data and provide datasets. Our work takes inspirations from all these previous works and we combine syllogism rules to develop synthetic data. Specially, we provide not only a dataset, but also a pipline that can give scores on LLM responses. Therefore, our method is suitable for on-policy training such as reward model and PPO training.

Many works improve reasoning ability from algorithmic aspect. \cite{wei2022chain} use Chain-of-Thought (CoT) method to improve reasoning ability, while \cite{wang2024chain} try some decoding method. \cite{havrilla2024teaching} and \cite{pang2024iterative} explore improving reasoning ability of LLM through RLHF and iterative DPO respectively. \cite{kumar2024training} attempt to incentive the self-correction ability of LLM. Our method mainly concentrates on RLHF method with our generated dataset. We show that data generation process has large performance effect on RLHF method.

\section{Preliminary}
\label{sec:preliminary}


\subsection{Proposition}


The proposition provides an object that we can judge to be true or false. Logical reasoning focuses on the process of deriving a conclusion from one or more premises. These premises and conclusions often appear in the form of propositions. In natural language, the expression of a proposition may be complex and diverse and it is hard to deal with such flexible data. In this article, we mainly focus on the syllogistic deduction in formal logic, which has relatively simple structure. Therefore, we mainly pay attention to the categorical propositions that judge the relationship between the subject and predicate. In categorical propositions, we judge the relationship between members of one category (the subject of the proposition) and another category (the object of the proposition). Aristotle has established four standard forms of categorical propositions, which we generally write as A, E, I, O~\citep{aristotle2021prior,copi2016introduction}. Below we use S as the subject and P as the predicate, and we give the formats of the four forms of propositions: (1) form \textit{A}: All S are P; (2) form \textit{E}: No S are P; (3) Form \textit{I}: There is one S that is P; (4) form \textit{O}: There is one S that is not P.

The four propositional structures presented here provide fundamental statements for judging the relationship between two entity categories. Such structures are independent of the actual semantics of the entities, allowing us to focus more easily on the formal relationships. For large-scale model data production, based on such formal structures, we can effectively carry out data production.

\subsection{Syllogism}
\label{sec:syllogism}

Herein, we briefly introduce the categorical syllogism in formal logic~\citep{copi2016introduction}. The data generation process in our subsequent work is predicated on these effective deductive reasoning rules.

The categorical syllogism is an ancient method of logical argumentation that employs deductive reasoning to study the inferential process of deriving a conclusion from two premises. In the syllogism, we refer to the subject of the conclusion as the minor term, and the predicate as the major term. The two premises of the syllogism each contain the major and minor terms, along with a term that connects the two premises, which we call the middle term. The premise containing the major term is referred to as the major premise, while the premise containing the minor term is called the minor premise. For instance, we may cite a classic example of a syllogism:
\begin{quote}
    ``All men are mortal. Socrates is a man. Therefore, Socrates is mortal.''
\end{quote}
Here, the major term is 'mortal', the minor term is 'Socrates', and the middle term is 'men'. Hence, "All men are mortal" is the major premise, and "Socrates is a man" is the minor premise.

The inference process of a categorical syllogism involves establishing a relationship between the major and minor terms using the middle term. In a valid categorical syllogism, if both the major and minor premises are true, then the conclusion must be true. Formal logic has provided several valid syllogistic forms. In our work, we utilize the 15 valid syllogistic forms\footnote{These valid syllogistic forms can be found in the table of all syllogisms in the wiki website: https://en.wikipedia.org/wiki/Syllogism. We only use results in solid line boxes.} found in modern formal logic \citep{copi2016introduction} to generate our logical data.

\section{MuseD Method}
\label{sec:mused_method}


Given our knowledge of valid syllogistic forms, we can in turn generate two premises from a conclusion by introducing a middle term. Conversely, in the actual deductive process, the relationship between the major and minor terms is established by eliminating the middle term. 
Therefore we can generate prompts in different complexity using the premises generation method. Moreover, we can score the responses based on the rules of deductive reasoning, thereby obtaining densely rewarded data. Additionally, our dataset can also serve as an evaluation set, allowing for the assessment of a model's deductive reasoning capabilities at various levels of complexity.

In the following, we will introduce the entire data generation process in four parts: prompt generation, response generation, response scoring and preference-pair composition.

\subsection{Prompt Generation}


The prompt for logical problems should be rigorous and reasonable. We need to ensure that the prompts we generate can lead to correct conclusions through multi-step reasoning. Therefore, we must guarantee correctness during the prompt generation process. Our approach to constructing prompts is backward-generated, starting from the conclusion to be proven and progressively increasing the complexity of the prompt based on valid formal rules of inference.


Our generation process consists of three steps. The first step is to create a deductive reasoning logic tree where each node on this tree is a proposition. The root node is the conclusion and the leaf nodes are the conditions in prompts. The generated tree only gives a logical structure where all terms are not filled. In the second step, we fill in the content for the missing entities in this tree. Finally, we combine conditions and conclusions to get prompts.


Specifically, we initially employ the syllogistic method of formal logic to generate a multi-step logic tree from the ultimate target conclusion. At the beginning of the tree generation process, we create a root node, which takes the form of one of the four propositions randomly sampled from the previous context. This root node is the final conclusion of this prompt. Throughout the process, we maintain a set of leaf nodes, which initially contains only the root node. As we progress, we iteratively pop a node from the set of leaf nodes, use it as a conclusion, sample a valid categorical syllogism pattern, and transform this conclusion node into two premise propositions by introducing a middle term. These two premise propositions are then added to the set of leaf nodes. Each such sampling process increases the number of leaf nodes by 1 and also necessitates an additional step in the multi-step reasoning from the leaf nodes back to the root node. Thus, by setting the number of times we expand the leaf nodes, we can control the complexity of the generated prompt. When we ultimately have $N+1$ leaf nodes, we turn them to $N+1$ propositions to construct the prompt. From these propositions, the final conclusion can be inferred through $N$ steps of reasoning. Note that during the generation process, we use placeholders rather than specific entities as terms to generate a logical framework. 


Based on the aforementioned logical conditions, we can also randomly introduce some interfering conditions. We introduce extra conditions that are related to the existing terms but do not conflict with current propositions, serving as noisy conditions in the reasoning process. 

After establishing the logical trees, we then filling placeholders with entities. The most straightforward yet challenging approach is to populate it with noun concepts that conform to the logical conditions. However, this method can easily lead models to take shortcuts \citep{zhang2023paradox,PrOntoQA} in concluding judgments rather than genuinely engaging in logical reasoning. For instance, given the premises that cats are mammals and mammals are animals, the model can easily arrive at this conclusion that cats are animals with common sense judgment rather than using deductive reasoning. This is what we aim to avoid.


In our work, we focus on the formal structure of logical reasoning itself. Therefore, we have chosen two forms of virtual entities: Greek letter names and virtual nouns. Greek letter names are directly sampled, such as ``ALPHA" or ``BETA", to substitute for entities. Virtual nouns consist of 4-14 English letters and are added to our virtual noun database after confirming they are not actual words. We then sample the required quantity from this database to fill the prompts.


Finally, we transform the conditions and conclusion with added entities into a prompt. For the conditions, we concatenate them directly as the given premises. Regarding the conclusion, we have two questioning formats: proof and judgment. For proof questions, we directly ask the model to prove the conclusion. For judgment questions, we reverse the conclusion to its negation with a probability of 0.5 and ask the model to judge whether the given proposition is correct. The templates for proof and judgement are given in Appendix~\ref{app:prompt_template}.

\subsection{Response Generation}





Based on the generated prompts, we can directly access the model to obtain responses. We refer to this method as ``natural" response generation. This approach yields the model's reasoning process expressed in natural language; however, such responses are not easily evaluated. We also incorporate instructions to generate responses in a specified JSON format within the prompt, allowing for structured answers that can be conveniently scored. We term this approach as ``formatted" response generation.

The natural method applies solely to the prompt itself, without additional requirements for the format and style of the response. We believe this is more consistent with the model's output during user interaction and aligns with the distribution of natural language. However, such natural language is relatively challenging to handle during scoring, necessitating additional operations in the evaluation phase to achieve better scoring outcomes.

The formatted generation method, on the other hand, introduces a fixed format for responses, resulting in outputs that are easier to score. Nevertheless, the significant difference between such structured expressions and natural language may impact the learning process during model alignment. The added format is provided in a few-shot manner, with details outlined in Appendix~\ref{app:format_template}.

We sample responses to the generated prompts using both methods. In our experiments, we utilize the Llama3 8b model and evaluate the model's responses under these two approaches.

\subsection{Response Evaluation}

The construction of formal logic problems are generated from the root node to the leaf node with the syllogism approach. Consequently, the procedure of proof or judgement constitutes the process of initiating from the leaf nodse and to arrive at the proposition of the root node.

We assess the responses of LLMs on formal logic reasoning problems in multiple aspects:

\begin{itemize}[leftmargin=*]
    \item \textbf{Step score}: Calculate the how much correct step the response reaches. We mainly calculate this score by counting the eliminated middle terms, details given in Sec.~\ref{sec:step_score}.
    \item \textbf{Result score}: Evaluate whether the response reaches the correct conclusion. A score of 0/1 is assigned. For proof-type questions, if the proposition to be proved emerges in the reasoning process, a score of 1 point is awarded. For judgment-type questions, the score is directly allocated based on the correctness of the judgment.
    \item \textbf{Intent score}: Evaluate whether the generated formatted response is valid JSON string. If valid, a score of 1 is given; otherwise, a score of 0 is assigned.
    \item \textbf{Wrong step}: Count the number of wrong reason steps. The scores equals the number of wrong steps in the propositions within the responses.
    \item \textbf{Noise step}: There might be propositions in the responses that we cannot check its correctness or irrelevant to the reasoning process. We count the numbers of these steps as the noise step score.
    \item \textbf{Extra step}: For correct steps in the responses, we count the number of repeated steps.
\end{itemize}

Step score, Result score, and Intent score are positive indicators. The higher the value, the better the effect of the LLM. Wrong step score, Noise step score, and Extra step score are negative indicators. The lower the value, the better the effect of the model.

\subsubsection{Step Score Calculation}
\label{sec:step_score}

For multi-step reasoning processes, we aim to obtain accurate process scores to evaluate the quality of different answers. It should be noted that the reason process from premises to the final conclusion is not unique. For example, consider conditions: (1) A is B, (2) B is C, and (3) C is D, and we want to conclude that (6) A is D. Using syllogistic reasoning, we can first derive (4) A is C from (1) and (2), and then combine condition (3) to reach (6). Alternatively, we can derive (5) B is D from (2) and (3), and then combine condition (1) to reach the conclusion (6). Both methods are correct and should receive equally high step scores.

Recall that the process of expanding logic tree is to add middle terms to get conditions. Hence reasoning process towards the final conclusion is essentially the process of eliminating these middle terms. In the above example, there are two paths to conduct inference. The first path eliminating B first and then C, while the second path eliminating C first and then B.

Based on observation above, we can enumerate valid middle terms in the logical tree, i.e. term B and C in the above example. We can check the reasoning process of each response and count the number of middle terms that are eliminated. Repeated elimination would not be counted. We calculate the rate of the eliminated middle terms and the number of all middle terms to get the step score.

\subsection{Preference-pair Composition}

We need to construct preference data to train the reward model (RM), based on the scored data above. Different from Outcome Reward Model (ORM)~\citep{lightmanlet}, we have multiple scoring dimensions to construct pairs. Therefore our RM can learn to provide dense reward signals to help PPO training. On the other hand, we don't train RM in the Process Reward Model (PRM) pattern~\cite{lightmanlet,Wang2023MathShepherdVA}, where rewards are assigned to each step, thus we can train RM with other datasets to ensure the general ability of RM.

Since each data point has multiple scoring dimensions, we can construct dataset with different composition methods, ultimately affecting the learning outcome of the model. Here, we experiment with three methods for generating data pairs:
\begin{itemize}[leftmargin=*]
    \item \textbf{P}: uses only positive scores (i.e., the step score and the result score) to construct pairs. For each pair, the chosen response must have a higher step score than the rejected one, while the result score of the chosen cannot be lower than that of the rejected one.
    \item \textbf{PN}: uses both positive and negative scores to construct pairs. We use all the scores above. For wrong, noise and extra scores, we use the opposite number to turn the negative signal into positive. For each pair, each score of the chosen response cannot be lower that that of the rejected one and the chosen should has at least one score that is higher.
    \item \textbf{R}: uses only result score as the pair standard. That is, the chosen response should has result score 1 while the rejected one has result score 0.
    \label{sec:rm_pair_method}
\end{itemize}

One important concern is that whether negative signals should be used. Intuitively, since it should be easier not to do wrong than to do right, negative signals may be easier to learn than positive ones. We compare method \textbf{P} and \textbf{PN} to see whether the choice of signals matters much. We also conduct method \textbf{R} to show the importance of step scores, since \textbf{R} only provides sparse final result signal, as usually used on ORM training.

\section{Experiments}
\label{sec:experiments}

\subsection{Evaluation Sets}
\label{sec:eval_sets}

We use three kinds of evaluation sets to test the performance of models. The in-domain (ID) set MuseD is generated and evaluated by our proposed method. The out-of-domain (OOD) sets are open source logical sets to show whether our model indeed improve its deductive logical ability on prompts out of distribution. We finally choose some general open source datasets to test whether the general abilities are influenced. We set the sampling temperature to be 0.3 for MuseD, 0.001 for OOD datasets.

\begin{itemize}[leftmargin=*]
    \item \textbf{ID set}: MuseD dataset with 2000 prompts, which is introduced in Sec.~\ref{sec:evaluation_mused}.
    \item \textbf{OOD sets}: For the open source logical verification set, we choose \textbf{PrOntoQA}~\citep{PrOntoQA}, \textbf{ProofWriter}~\citep{tafjord2021proofwriter}, \textbf{LogicalDeduction}~\citep{pan2023logiclmempoweringlargelanguage}, \textbf{FOLIO}~\citep{han2024folionaturallanguagereasoning}, \textbf{AR-LSAT}~\citep{zhong2021arlsatinvestigatinganalyticalreasoning}. 
    We use the CoT (Chain of Thought)~\citep{wei2022chain} method, and CoT prompt of each dataset from ~\citep{xu2024faithful}. The evaluation metric is accuracy, which measures the correctness of multiple-choice questions.
    \item \textbf{General ability sets}
    We follow the default implementation setting of LM Eval Harness \citep{eval-harness} and set the temperature hyperparameter as 0.
    For general ability evaluation, we select the task evaluation in Sec.~\ref{sec:common_evaluate} and follow the default implementation setting of LM Eval Harness with temperature hyperparameter as 0 and report 0-shot accuracy.
\end{itemize}


\subsection{Reward Model}
\label{sec:rm}

The reward models (RM) are trained following the standard process of InstructGPT~\citep{ouyang2022training}. We choose the Llama3 \citep{dubey2024llama} 8B pretrain model as our base model. If not specified, we keep the hyper-parameters of RMs fixed. More specifically, we set the batchsize to be 384 and the maximum learning rate to be 5e-6.

We train RMs with different preference datasets. We mainly to evaluate the performance of our constructed logic dataset. We choose UltraFeedback (UF)~\citep{cui2023ultrafeedback} as our base dataset to avoid ability decline in other fields, we use the fine-grained scores to get around UF 27w pairs. For our logic dataset, we compare the performances of the natural and formatted response generation methods. We also test the P, PN and R pair construction methods. All our RMs are listed below. For convenience, we use \textbf{Na} to denote natural responses and \textbf{Fo} to denote formatted responses.

\begin{itemize}[leftmargin=*]
    \item \textbf{RM-UF}: uses only 27w UF data.
    \item \textbf{RM-Na-P}: uses 15k natural P-pair logic data. UF data is used.
    \item \textbf{RM-Fo-P}: uses 15k formatted P-pair logic data. UF data is used.
    \item \textbf{RM-Mix-P}: uses 16k logic data, half from the logic data in RM-Na-P and half from the data in RM-Fo-P. UF data is used.
    \item \textbf{RM-Na-PN}: uses 11k natural PN-pair logic data. UF data is used.
    \item \textbf{RM-Na-R}: uses 8k natural R-pair logic data. UF data is used.
    \item \textbf{RM-NaO-P}: uses only 15k natural P-pair logic data. We turn the batchsize of this RM to be 96 to get more steps.
\end{itemize}

Notice that we use the same natural responses to construct P, PN and R pairs and different rules result in different size of dataset. For Fo-P data, we downsample them to the same size of Na-P data. We put the validation accuracy and RewardBench~\citep{lambert2024rewardbench} scores of these RMs in the Appendix~\ref{append:rm_acc}. Roughly speaking, each RM gains high validation accuracy on its trained datasets, and the reasoning scores in RewardBench are not directly consistent with our experimental results in Sec.~\ref{sec:experiments}, possibly due to the distribution mismatch.

\subsection{PPO Training}
\subsubsection{Settings}

We train PPO models using above RMs. The base model for PPO training is Llama3 8B Instruct model. The prompts used for training are the same prompts used for the corresponding RM. That is, the prompts used for one RM would be used for the PPO model that uses this RM. The hyperparameters for training PPO are fixed for all experiments. The experience batch size is 160 and the micro batch size is 20. The sampling temperature for training is 1.0. We train for only one epoch. For convenience, we denote each PPO model with its corresponding RM as its subscript. We remove the $\mathrm{RM-}$ in the subscript for briefly. That is, we use $\mathrm{PPO}_{\mathrm{A}}$ to denote the PPO trained with RM-A.



Further, we compare the performance difference if we change the training prompts of PPO. We choose RM-Na-P as our baseline RM and use $\mathrm{PPO}_{\mathrm{Na-P}}$ as our baseline PPO model. We test the impact of using extra in-domain prompts. For $\mathrm{PPO}_{\mathrm{Na-P}}$, we use 6587 prompts trained for RM-Na-P. Then construct more prompts with our pipline MuseD to increase the logical prompts to reach 20000. We combine all these prompts with UF prompts to train one epoch over PPO training, and we get $\mathrm{PPO}_{\mathrm{Na-P-All}}$. We also sample 6587 prompts from the 20000 prompts to train PPO and we get $\mathrm{PPO}_{\mathrm{Na-P-Replace}}$. Finally, we test the effect of curriculum learning during PPO training. That is, we sort the logical prompts used in $\mathrm{PPO}_{\mathrm{Na-P}}$ with their levels. We train them from easy to hard and we get $\mathrm{PPO}_{\mathrm{Na-P-Cur}}$.

\subsubsection{Performances}


We compare the performances of all PPO models on the three kinds of evaluation sets mentioned in Sec. \ref{sec:eval_sets}. The performances of all models on our MuseD dataset, i.e. the ID dataset, are shown in Table~\ref{tab:all_self_logic}. Table~\ref{tab:all_open_logit_eval} gives comparison of models on the OOD datasets. The results for general ability of models are shown in Appendix \ref{append:General_set}. Compared with $\mathrm{PPO}_{\mathrm{UF}}$, the involvement of our data do not seem to cause degradation on other abilities.

Compared to $\mathrm{PPO}_{\mathrm{Na-P}}$, $\mathrm{PPO}_{\mathrm{NaO-P}}$ shows bad performance on all scores, indicating the necessity of using a general dataset. Compared to Llama3 8B Instruct, the UF dataset only don't increase the step and result scores much, while on out-of-domain datasets, $\mathrm{PPO}_{\mathrm{UF}}$ has a obvious improvement. We check the cases and find that the bad performance on instruction following leads to the low scores of Llama3-8b-Instruct. It can be seen from the results on AR-LSAT, which is a 5-choice dataset. A random strategy would leads to a score around 0.2 while Llama3-8b-Instruct gets a score near 0.

To show the effect of training with our dataset, we compare $\mathrm{PPO}_{\mathrm{UF}}$ and $\mathrm{PPO}_{\mathrm{Na-P}}$. The logical data yields a significant gain in logical effects. Compared to $\mathrm{PPO}_{\mathrm{UF}}$, $\mathrm{PPO}_{\mathrm{Na-P}}$ increases the step and result scores of MuseD both by 12 percentage points, and reduces the average wrong step significantly. The noise step and extra step indicators are basically the same. More results on level-wise comparison are given in Appendix~\ref{appendix:detail_comparison}. On the public evaluation datasets, there is significant growth on ProntoQA, ProofWriter, and LogicalDeduction. Specifically, there are 14-16\% improvements on ProntoQA and LogicalDeduction, which have quite similar deductive tasks as MuseD dataset. There is a slightly decrease on AR-LAST, a five-choice task dataset (refer to Appendix~\ref{app:ar_lsat_demo}). Although both PPO models get a reward similar to a random strategy, $\mathrm{PPO}_{\mathrm{Na-P}}$ sometimes refused to pick any answer when it thinks all options are wrong.

\begin{table}[H]
\centering
\setlength\tabcolsep{5pt}
\adjustbox{max width=\linewidth}{
\begin{tabular}{@{}l|ccc|ccccccc@{}}
\toprule
\textbf{Model} & \textbf{step score} & \textbf{result score} & \textbf{intent score} & \textbf{wrong step count} & \textbf{noise step count} & \textbf{extra step count} \\ \midrule
Llama3-8B-Instruct & 0.3485 & 0.5715 & 0.9495 & 1.685 & 0.7815 & 0.054 \\ 
$\mathrm{PPO}_{\mathrm{UF}}$ & 0.3135 & 0.5725 & 0.9905 & 1.252 & 0.438 & 0.0305 \\ \midrule
$\mathrm{PPO}_{\mathrm{Na-P}}$ & 0.4383 & 0.6975 & 0.9925 & 0.85 & 0.3865 & 0.036 \\
$\mathrm{PPO}_{\mathrm{NaO-P}}$ & 0.4117 & 0.4920 & 0.8695 & 2.3715 & 1.0515 & 0.5585 \\ 
$\mathrm{PPO}_{\mathrm{Na-P-All}}$ & \textbf{0.4993} & \textbf{0.7585} & 0.9905 & 0.6645 & 0.3595 & 0.0805 \\ 
$\mathrm{PPO}_{\mathrm{Na-P-Replace}}$ & 0.431 & 0.676 & 0.985 & 0.842 & 0.425 & 0.05 \\ 
$\mathrm{PPO}_{\mathrm{Na-P-Cur}}$ & 0.4591 & 0.6855 & 0.991 & 0.9515 & 0.455 & 0.0655 \\ \midrule
$\mathrm{PPO}_{\mathrm{Na-R}}$ & 0.3472 & 0.6605 & \textbf{0.9975} & 0.8115 & 0.257 & 0.018 \\
$\mathrm{PPO}_{\mathrm{Na-PN}}$ & 0.3649 & 0.6675 & 0.99 & \textbf{0.672} & \textbf{0.2155} & \textbf{0.0125} \\ \midrule
$\mathrm{PPO}_{\mathrm{Fo-P}}$ & 0.3744 & 0.6205 & 0.9725 & 1.0945 & 0.367 & 0.053 \\
$\mathrm{PPO}_{\mathrm{Mix-P}}$ & 0.4455 & 0.693 & 0.993 & 1.135 & 0.5615 & 0.0645 & \\
\bottomrule
\end{tabular}}
    \caption{Performances of Lamma3 8B models trained with different RMs on MuseD dataset.}
    \label{tab:all_self_logic}
\end{table}

\begin{table}[H]
\centering
\setlength\tabcolsep{5pt}
\adjustbox{max width=\linewidth}{
\begin{tabular}{@{}l|ccccc|ccccc@{}}
\toprule
\textbf{Model} & \textbf{ProntoQA} & \textbf{ProofWriter} & \textbf{LogicalDeduction} & \textbf{FOLIO} & \textbf{AR-LSAT} & \textbf{Average} \\ \midrule
Llama3-8B-Instruct & 0.616 & 0.1367 & 0.0567 & 0.1813 & 0.0043 & 0.199 \\ 
$\mathrm{PPO}_{\mathrm{UF}}$ & 0.728 & 0.37 & 0.3267 & 0.4118 & 0.2165 & 0.411 \\ \midrule
$\mathrm{PPO}_{\mathrm{Na-P}}$ & \textbf{0.868} & 0.4567 & \textbf{0.4833} & 0.451 & 0.1948 & \textbf{0.4908} \\
$\mathrm{PPO}_{\mathrm{NaO-P}}$ & 0.352 & 0.2467 & 0.0967 & 0.2451 & 0.0519 & 0.1985 \\
$\mathrm{PPO}_{\mathrm{Na-P-All}}$ & 0.672 & 0.4517 & 0.32 & \textbf{0.4804} & 0.1905 & 0.4229\\
$\mathrm{PPO}_{\mathrm{Na-P-Replace}}$ & 0.818 & \textbf{0.4967} & 0.3933 & 0.4803 & 0.2035 & 0.4784\\
$\mathrm{PPO}_{\mathrm{Na-P-Cur}}$ & 0.762 & 0.4717 & 0.4066 & 0.4706 & 0.1861 & 0.4594 \\ \midrule
$\mathrm{PPO}_{\mathrm{Na-R}}$ & 0.798 & 0.4017 & 0.3867 & 0.4069 & \textbf{0.2294} & 0.4445 \\
$\mathrm{PPO}_{\mathrm{Na-PN}}$ & 0.816 & 0.445 & 0.2667 & 0.402 & 0.1948 & 0.4249 \\ \midrule
$\mathrm{PPO}_{\mathrm{Fo-P}}$ & 0.804 & 0.4033 & 0.44 & 0.4461 & 0.1905 & 0.4568 \\
$\mathrm{PPO}_{\mathrm{Mix-P}}$ & 0.806 & 0.4367 & 0.43 & 0.4265 & 0.1991 & 0.4597 \\
\bottomrule
\end{tabular}}
    \caption{Performances of Lamma3 8B models trained with different RMs on out-domain logical evaluation sets.}
    \label{tab:all_open_logit_eval}
\end{table}

\subsubsection{Preference Pair Composition}


We compare the performance of models trained with PPO across different preference pair generation methods which are presented in Section ~\ref{sec:rm_pair_method}. We compare the performances of $\mathrm{PPO}_{\mathrm{Na-P}}$, $\mathrm{PPO}_{\mathrm{Na-PN}}$ and $\mathrm{PPO}_{\mathrm{Na-R}}$.

\textbf{Process signals can significantly improve the effect.} RM-Na-P uses preference data incorporating process signals while RM-Na-R uses data constructed with result score only. The step score exhibits significant growth on the MuseD evaluation set. On the public datasets, $\mathrm{PPO}_{\mathrm{Na-P}}$ is notably superior to $\mathrm{PPO}_{\mathrm{Na-R}}$. There are about 10-point improvement LogicalDeduction, 7-point imporvement on ProntoQA , and 5-point improvements on both ProofWriter and FOLIO . 

\textbf{Negative signals will cause degradation in effectiveness.} Introducing negative signals for the construction of preference data in the reward model will notably diminish the effectiveness of the PPO model. $\mathrm{PPO}_{\mathrm{Na-PN}}$ further incorporates negative indicators on the foundation of $\mathrm{PPO}_{\mathrm{Na-P}}$. There is indeed a reduction in the three dimensions of negative indicators (wrong step count, noise step count, and extra step count) in MuseD. However, it significantly deteriorates in the positive indicator step score. Concurrently, on the public logical verification set, ProntoQA, LogicalDeduction, and FOLIO exhibit a significant decline.

\subsubsection{Natural response VS. Formatted response}

We also investigate the influence of data format of prompts. We choose Na (natural responses) and Fo (formatted responses). We train three PPO models with distinct data formats: $\mathrm{PPO}_{\mathrm{Na-P}}$ (utilizing Na), $\mathrm{PPO}_{\mathrm{Fo-P}}$ (employing Fo), and $\mathrm{PPO}_{\mathrm{Mix-P}}$ (mixing Na and Fo in a 1:1 ratio).

\textbf{Natural format performs better.} $\mathrm{PPO}_{\mathrm{Na-P}}$ is notably superior to $\mathrm{PPO}_{\mathrm{Fo-P}}$ and $\mathrm{PPO}_{\mathrm{Mix-P}}$. On MuseD dataset, $\mathrm{PPO}_{\mathrm{Na-P}}$ exhibits significant superiority over $\mathrm{PPO}_{\mathrm{Fo-P}}$, boasting a 6-7 point advantage in step score and result score. In comparison to $\mathrm{PPO}_{\mathrm{Mix-P}}$ version, $\mathrm{PPO}_{\mathrm{Na-P}}$ version is generally on an equal footing in positive indicators. However, regarding negative indicators such as incorrect step count, noise step count, and extra step count, $\mathrm{PPO}_{\mathrm{Na-P}}$ version is markedly better than $\mathrm{PPO}_{\mathrm{Mix-P}}$ version. On the public test logic test collection, $\mathrm{PPO}_{\mathrm{Na-P}}$ version surpasses $\mathrm{PPO}_{\mathrm{Fo-P}}$ and $\mathrm{PPO}_{\mathrm{Mix-P}}$ version in the ProntoQA, ProofWriter, LogicalDeduction, and FOLIO datasets. The result shows that natural format, which is more closely to the natural language, is more effective for RLHF training.


\subsubsection{Variants of logical prompts in PPO training}
We also experiment on the impact of using additional prompts on top of the prompts covered by RM training. We compare the performances among $\mathrm{PPO}_{\mathrm{Na-P}}$, $\mathrm{PPO}_{\mathrm{Na-P-All}}$ and $\mathrm{PPO}_{\mathrm{Na-P-Replace}}$.


Compare $\mathrm{PPO}_{\mathrm{Na-P-All}}$ with $\mathrm{PPO}_{\mathrm{Na-P}}$. It can be observed that by adding in-domain prompts, the effect on the MuseD set is significantly increased, while there is a significant drop on the out-of-domain data set. It seems that RM-Na-P has the ability to get relatively correct scores for these in-domain prompts, which helps PPO model to improve. However, a too high data proportion will cause overfittin on MuseD data and reduce the effect on non-identically distributed datasets.

Compare the two experiments of $\mathrm{PPO}_{\mathrm{Na-P-Replace}}$ and $\mathrm{PPO}_{\mathrm{Na-P}}$. In the MuseD set, $\mathrm{PPO}_{\mathrm{Na-P}}$ exhibits superiority in all indicators to $\mathrm{PPO}_{\mathrm{Na-P-Replace}}$. On the publicly available logic test set, $\mathrm{PPO}_{\mathrm{Na-P-Replace}}$ shows worse performance on ProntoQA and LogicalDeduction but better performance on FOLIO and ProofWriter. On average, it decreased by approximately 1 point. Generally, using prompts trained with the reward model to train PPO yields better effects both on in-distribution prompts and on open-source data outside the distribution. The possible reason is that on the trained prompts, the reward model scores more accurately, thereby better guiding the training process of PPO. As for the better performance on FOLIO and ProofWriter, it might be caused by a bit overfitting on the MuseD deductive format, which hurts the reasoning ability on tasks like FOLIO or ProofWriter a bit, since ProntoQA and LogicalDeduction have more similar formats to MuseD.

\subsubsection{curriculum learning}
We also conducte an experiment on curriculum learning as $\mathrm{PPO}_{\mathrm{Na-P-cur}}$ to study its impact on the effect. We uniformly incorporate data of varying levels, ranging from low to high, into the prompt of PPO training. Ensure that with PPO steps training, the traversed logical data progresses from simple to complex. 

It can be observed that in comparison to shuffling data, employing the mechanism of curriculum learning: on the MuseD set, the overall result is comparable. However, on the open-source test set, the performance drops. On ProntoQA and LogicalDeduction, it declines by 8 to 10 percentage points.

\section{Evaluations on LLMs}
\label{sec:evaluation_mused}
Our constructed data can also be used as an evaluation dataset to test the performance of LLMs on multi-step deductive reasoning. We generate 2000 prompts to compose this evaluation sets, which we called MuseD. MuseD includes questions that multi-step deductive reasoning is needed. The reason steps range from 1 to 10. 

We test the effects of some models on MuseD and the results are shown in Table \ref{tab:mused_eval}. Obviously, GPT-4-o1-Preview shows best performance. Qwen2.5-72B-Instruct also reaches high scores on step and result scores.
\begin{table}[H]
\centering
\setlength\tabcolsep{5pt}
\adjustbox{max width=\linewidth}{
\begin{tabular}{@{}lcccccccccc@{}}
\toprule
\textbf{Model} & \textbf{step score} & \textbf{result score} & \textbf{intent score} & \textbf{wrong step count} & \textbf{noise step count} & \textbf{extra step count} \\ \midrule

GPT-4 & 0.7306 & 0.8250 & 0.9995 & 1.1015 & 0.4785 & \textbf{0.0125}  \\
GPT-4o & 0.8085 & 0.8320 & 0.9890 & 0.6940 & 0.7490 & 0.1015  \\
GPT-4o-mini & 0.4514 & 0.6635 & \textbf{1.0000} & 1.3570 & 0.6550 & 0.0290  \\
GPT-4-o1-preview & \textbf{0.8516} & \textbf{0.8895} & 0.9785 & \textbf{0.2785} & \textbf{0.0380} & 0.0800  \\
GPT-4-o1-mini & 0.7619 & 0.8745 & 0.9980 & 0.6210 & 0.0865 & 0.0430 \\
Qwen2.5-72B-Instruct & 0.8236 & 0.8844 & 0.9995 & 0.6386 & 0.5796 & 0.0581  \\
Qwen2-72B-Instruct & 0.7316 & 0.8050 & 0.9980 & 1.2130 & 0.9310 & 0.0675 \\
Llama3.1-72B-Instruct & 0.7543 & 0.8370 & 0.9920 & 1.3005 & 0.8135 & 0.1020 \\

\bottomrule
\end{tabular}}
    \caption{Performances of Different Models on MuseD.}
    \label{tab:mused_eval}
\end{table}

Using our MuseD evaluation set, we can see the detailed performances of LLMs. Besides the positive and negative scores above, we show the performance of LLMs over different levels and noise counts, as shown in Appendix~\ref{app:detailed_eval}.

\section{Conclusion}

In this work, we propose a multi-step deductive data generation pipline, MuseD, including prompt generation, response scoring and pair composition. MuseD can construct prompts with controllable complexity and check the step scores of responses. We validate the effect of our generated logical data on Lamma3 8B instruct model with RLHF method. The result shows that our data can lead to significant improvement on in-domain of out-of-domain deductive reasoning tasks. Further, we show that natural format and positive step signals are important for RLHF. Finally, we use our pipline to generate an evaluation dataset, also named MuseD, to evaluate the performance of current LLMs.

\bibliography{iclr2025_conference}

\begin{thebibliography}{36}
\providecommand{\natexlab}[1]{#1}
\providecommand{\url}[1]{\texttt{#1}}
\expandafter\ifx\csname urlstyle\endcsname\relax
  \providecommand{\doi}[1]{doi: #1}\else
  \providecommand{\doi}{doi: \begingroup \urlstyle{rm}\Url}\fi

\bibitem[Aristotle(350 BCE)]{aristotle2021prior}
Aristotle.
\newblock Prior analytics.
\newblock 350 BCE.

\bibitem[Bai et~al.(2022)Bai, Jones, Ndousse, Askell, Chen, Dassarma, Drain, Fort, Ganguli, and Henighan]{2022Training}
Yuntao Bai, Andy Jones, Kamal Ndousse, Amanda Askell, Anna Chen, Nova Dassarma, Dawn Drain, Stanislav Fort, Deep Ganguli, and Tom Henighan.
\newblock Training a helpful and harmless assistant with reinforcement learning from human feedback.
\newblock 2022.

\bibitem[Bhagavatula et~al.()Bhagavatula, Le~Bras, Malaviya, Sakaguchi, Holtzman, Rashkin, Downey, Yih, and Choi]{bhagavatulaabductive}
Chandra Bhagavatula, Ronan Le~Bras, Chaitanya Malaviya, Keisuke Sakaguchi, Ari Holtzman, Hannah Rashkin, Doug Downey, Wen-tau Yih, and Yejin Choi.
\newblock Abductive commonsense reasoning.
\newblock In \emph{International Conference on Learning Representations}.

\bibitem[Clark et~al.(2018)Clark, Cowhey, Etzioni, Khot, Sabharwal, Schoenick, and Tafjord]{Clark2018ThinkYH}
Peter Clark, Isaac Cowhey, Oren Etzioni, Tushar Khot, Ashish Sabharwal, Carissa Schoenick, and Oyvind Tafjord.
\newblock Think you have solved question answering? try arc, the ai2 reasoning challenge.
\newblock \emph{ArXiv}, abs/1803.05457, 2018.

\bibitem[Cobbe et~al.(2021)Cobbe, Kosaraju, Bavarian, Hilton, Nakano, Hesse, and Schulman]{cobbe2021training}
Karl Cobbe, Vineet Kosaraju, Mohammad Bavarian, Jacob Hilton, Reiichiro Nakano, Christopher Hesse, and John Schulman.
\newblock Training verifiers to solve math word problems, 2021.

\bibitem[Copi et~al.(2016)Copi, Cohen, and McMahon]{copi2016introduction}
Irving~M Copi, Carl Cohen, and Kenneth McMahon.
\newblock \emph{Introduction to logic}.
\newblock Routledge, 2016.

\bibitem[Cui et~al.(2023)Cui, Yuan, Ding, Yao, Zhu, Ni, Xie, Liu, and Sun]{cui2023ultrafeedback}
Ganqu Cui, Lifan Yuan, Ning Ding, Guanming Yao, Wei Zhu, Yuan Ni, Guotong Xie, Zhiyuan Liu, and Maosong Sun.
\newblock Ultrafeedback: Boosting language models with high-quality feedback.
\newblock \emph{arXiv preprint arXiv:2310.01377}, 2023.

\bibitem[Dubey et~al.(2024)Dubey, Jauhri, Pandey, Kadian, Al-Dahle, Letman, Mathur, Schelten, Yang, Fan, et~al.]{dubey2024llama}
Abhimanyu Dubey, Abhinav Jauhri, Abhinav Pandey, Abhishek Kadian, Ahmad Al-Dahle, Aiesha Letman, Akhil Mathur, Alan Schelten, Amy Yang, Angela Fan, et~al.
\newblock The llama 3 herd of models.
\newblock \emph{arXiv preprint arXiv:2407.21783}, 2024.

\bibitem[Gao et~al.(2024)Gao, Tow, Abbasi, Biderman, Black, DiPofi, Foster, Golding, Hsu, Le~Noac'h, Li, McDonell, Muennighoff, Ociepa, Phang, Reynolds, Schoelkopf, Skowron, Sutawika, Tang, Thite, Wang, Wang, and Zou]{eval-harness}
Leo Gao, Jonathan Tow, Baber Abbasi, Stella Biderman, Sid Black, Anthony DiPofi, Charles Foster, Laurence Golding, Jeffrey Hsu, Alain Le~Noac'h, Haonan Li, Kyle McDonell, Niklas Muennighoff, Chris Ociepa, Jason Phang, Laria Reynolds, Hailey Schoelkopf, Aviya Skowron, Lintang Sutawika, Eric Tang, Anish Thite, Ben Wang, Kevin Wang, and Andy Zou.
\newblock A framework for few-shot language model evaluation, 07 2024.
\newblock URL \url{https://zenodo.org/records/12608602}.

\bibitem[Han et~al.(2024)Han, Schoelkopf, Zhao, Qi, Riddell, Zhou, Coady, Peng, Qiao, Benson, Sun, Wardle-Solano, Szabo, Zubova, Burtell, Fan, Liu, Wong, Sailor, Ni, Nan, Kasai, Yu, Zhang, Fabbri, Kryscinski, Yavuz, Liu, Lin, Joty, Zhou, Xiong, Ying, Cohan, and Radev]{han2024folionaturallanguagereasoning}
Simeng Han, Hailey Schoelkopf, Yilun Zhao, Zhenting Qi, Martin Riddell, Wenfei Zhou, James Coady, David Peng, Yujie Qiao, Luke Benson, Lucy Sun, Alex Wardle-Solano, Hannah Szabo, Ekaterina Zubova, Matthew Burtell, Jonathan Fan, Yixin Liu, Brian Wong, Malcolm Sailor, Ansong Ni, Linyong Nan, Jungo Kasai, Tao Yu, Rui Zhang, Alexander~R. Fabbri, Wojciech Kryscinski, Semih Yavuz, Ye~Liu, Xi~Victoria Lin, Shafiq Joty, Yingbo Zhou, Caiming Xiong, Rex Ying, Arman Cohan, and Dragomir Radev.
\newblock Folio: Natural language reasoning with first-order logic, 2024.
\newblock URL \url{https://arxiv.org/abs/2209.00840}.

\bibitem[Havrilla et~al.()Havrilla, Du, Raparthy, Nalmpantis, Dwivedi-Yu, Hambro, Sukhbaatar, and Raileanu]{havrilla2024teaching}
Alexander Havrilla, Yuqing Du, Sharath~Chandra Raparthy, Christoforos Nalmpantis, Jane Dwivedi-Yu, Eric Hambro, Sainbayar Sukhbaatar, and Roberta Raileanu.
\newblock Teaching large language models to reason with reinforcement learning.
\newblock In \emph{AI for Math Workshop@ ICML 2024}.

\bibitem[Hendrycks et~al.(2020)Hendrycks, Burns, Basart, Zou, Mazeika, Song, and Steinhardt]{Hendrycks_Burns_Basart_Zou_Mazeika_Song_Steinhardt_2020}
Dan Hendrycks, Collin Burns, Steven Basart, Andy Zou, Mantas Mazeika, Dawn Song, and Jacob Steinhardt.
\newblock Measuring massive multitask language understanding.
\newblock \emph{Cornell University - arXiv,Cornell University - arXiv}, Sep 2020.

\bibitem[Joshi et~al.(2017)Joshi, Choi, Weld, and Zettlemoyer]{JoshiTriviaQA2017}
Mandar Joshi, Eunsol Choi, Daniel~S. Weld, and Luke Zettlemoyer.
\newblock Triviaqa: A large scale distantly supervised challenge dataset for reading comprehension.
\newblock In \emph{Proceedings of the 55th Annual Meeting of the Association for Computational Linguistics}, Vancouver, Canada, July 2017. Association for Computational Linguistics.

\bibitem[Kumar et~al.(2024)Kumar, Zhuang, Agarwal, Su, Co-Reyes, Singh, Baumli, Iqbal, Bishop, Roelofs, et~al.]{kumar2024training}
Aviral Kumar, Vincent Zhuang, Rishabh Agarwal, Yi~Su, John~D Co-Reyes, Avi Singh, Kate Baumli, Shariq Iqbal, Colton Bishop, Rebecca Roelofs, et~al.
\newblock Training language models to self-correct via reinforcement learning.
\newblock \emph{arXiv preprint arXiv:2409.12917}, 2024.

\bibitem[Lambert et~al.(2024)Lambert, Pyatkin, Morrison, Miranda, Lin, Chandu, Dziri, Kumar, Zick, Choi, et~al.]{lambert2024rewardbench}
Nathan Lambert, Valentina Pyatkin, Jacob Morrison, LJ~Miranda, Bill~Yuchen Lin, Khyathi Chandu, Nouha Dziri, Sachin Kumar, Tom Zick, Yejin Choi, et~al.
\newblock Rewardbench: Evaluating reward models for language modeling.
\newblock \emph{arXiv preprint arXiv:2403.13787}, 2024.

\bibitem[Lightman et~al.()Lightman, Kosaraju, Burda, Edwards, Baker, Lee, Leike, Schulman, Sutskever, and Cobbe]{lightmanlet}
Hunter Lightman, Vineet Kosaraju, Yuri Burda, Harrison Edwards, Bowen Baker, Teddy Lee, Jan Leike, John Schulman, Ilya Sutskever, and Karl Cobbe.
\newblock Let's verify step by step.
\newblock In \emph{The Twelfth International Conference on Learning Representations}.

\bibitem[Lin et~al.(2022)Lin, Hilton, and Evans]{lin-etal-2022-truthfulqa}
Stephanie Lin, Jacob Hilton, and Owain Evans.
\newblock {T}ruthful{QA}: Measuring how models mimic human falsehoods.
\newblock In \emph{Proceedings of the 60th Annual Meeting of the Association for Computational Linguistics (Volume 1: Long Papers)}, pp.\  3214--3252, Dublin, Ireland, May 2022. Association for Computational Linguistics.
\newblock \doi{10.18653/v1/2022.acl-long.229}.
\newblock URL \url{https://aclanthology.org/2022.acl-long.229}.

\bibitem[Liu et~al.(2023)Liu, Liu, Cui, Teng, Duan, Zhou, and Zhang]{liu2023logiqa}
Hanmeng Liu, Jian Liu, Leyang Cui, Zhiyang Teng, Nan Duan, Ming Zhou, and Yue Zhang.
\newblock Logiqa 2.0—an improved dataset for logical reasoning in natural language understanding.
\newblock \emph{IEEE/ACM Transactions on Audio, Speech, and Language Processing}, 2023.

\bibitem[Liu et~al.(2021)Liu, Cui, Liu, Huang, Wang, and Zhang]{liu2020logiqa}
Jian Liu, Leyang Cui, Hanmeng Liu, Dandan Huang, Yile Wang, and Yue Zhang.
\newblock Logiqa: a challenge dataset for machine reading comprehension with logical reasoning.
\newblock In \emph{Proceedings of the Twenty-Ninth International Conference on International Joint Conferences on Artificial Intelligence}, pp.\  3622--3628, 2021.

\bibitem[Mihaylov et~al.(2018)Mihaylov, Clark, Khot, and Sabharwal]{OpenBookQA2018}
Todor Mihaylov, Peter Clark, Tushar Khot, and Ashish Sabharwal.
\newblock Can a suit of armor conduct electricity? a new dataset for open book question answering.
\newblock In \emph{EMNLP}, 2018.

\bibitem[Ouyang et~al.(2022)Ouyang, Wu, Jiang, Almeida, Wainwright, Mishkin, Zhang, Agarwal, Slama, Ray, et~al.]{ouyang2022training}
Long Ouyang, Jeffrey Wu, Xu~Jiang, Diogo Almeida, Carroll Wainwright, Pamela Mishkin, Chong Zhang, Sandhini Agarwal, Katarina Slama, Alex Ray, et~al.
\newblock Training language models to follow instructions with human feedback.
\newblock \emph{Advances in neural information processing systems}, 35:\penalty0 27730--27744, 2022.

\bibitem[Pan et~al.(2023)Pan, Albalak, Wang, and Wang]{pan2023logiclmempoweringlargelanguage}
Liangming Pan, Alon Albalak, Xinyi Wang, and William~Yang Wang.
\newblock Logic-lm: Empowering large language models with symbolic solvers for faithful logical reasoning, 2023.
\newblock URL \url{https://arxiv.org/abs/2305.12295}.

\bibitem[Pang et~al.(2024)Pang, Yuan, Cho, He, Sukhbaatar, and Weston]{pang2024iterative}
Richard~Yuanzhe Pang, Weizhe Yuan, Kyunghyun Cho, He~He, Sainbayar Sukhbaatar, and Jason Weston.
\newblock Iterative reasoning preference optimization.
\newblock \emph{arXiv preprint arXiv:2404.19733}, 2024.

\bibitem[Paperno et~al.(2016)Paperno, Kruszewski, Lazaridou, Pham, Bernardi, Pezzelle, Baroni, Boleda, and Fernández]{lambada_openai}
Denis Paperno, Germán Kruszewski, Angeliki Lazaridou, Quan~Ngoc Pham, Raffaella Bernardi, Sandro Pezzelle, Marco Baroni, Gemma Boleda, and Raquel Fernández.
\newblock The lambada dataset, Aug 2016.

\bibitem[Sanh et~al.(2022)Sanh, Webson, Raffel, Bach, Sutawika, Alyafeai, Chaffin, Stiegler, Le~Scao, Raja, et~al.]{sanh2022multitask}
Victor Sanh, Albert Webson, Colin Raffel, Stephen~H Bach, Lintang Sutawika, Zaid Alyafeai, Antoine Chaffin, Arnaud Stiegler, Teven Le~Scao, Arun Raja, et~al.
\newblock Multitask prompted training enables zero-shot task generalization.
\newblock In \emph{ICLR 2022-Tenth International Conference on Learning Representations}, 2022.

\bibitem[Saparov \& He(2023)Saparov and He]{PrOntoQA}
Abulhair Saparov and He~He.
\newblock Language models are greedy reasoners: A systematic formal analysis of chain-of-thought.
\newblock In \emph{The Eleventh International Conference on Learning Representations}, 2023.
\newblock URL \url{https://openreview.net/forum?id=qFVVBzXxR2V}.

\bibitem[Saparov et~al.(2024)Saparov, Pang, Padmakumar, Joshi, Kazemi, Kim, and He]{PrOntoQAOOD}
Abulhair Saparov, Richard~Yuanzhe Pang, Vishakh Padmakumar, Nitish Joshi, Mehran Kazemi, Najoung Kim, and He~He.
\newblock Testing the general deductive reasoning capacity of large language models using ood examples.
\newblock \emph{Advances in Neural Information Processing Systems}, 36, 2024.

\bibitem[Sun et~al.(2023)Sun, Zheng, Xie, Liu, Chu, Qiu, Xu, Ding, Li, Geng, et~al.]{sun2023survey}
Jiankai Sun, Chuanyang Zheng, Enze Xie, Zhengying Liu, Ruihang Chu, Jianing Qiu, Jiaqi Xu, Mingyu Ding, Hongyang Li, Mengzhe Geng, et~al.
\newblock A survey of reasoning with foundation models.
\newblock \emph{arXiv preprint arXiv:2312.11562}, 2023.

\bibitem[Tafjord et~al.(2021)Tafjord, Dalvi, and Clark]{tafjord2021proofwriter}
Oyvind Tafjord, Bhavana Dalvi, and Peter Clark.
\newblock Proofwriter: Generating implications, proofs, and abductive statements over natural language.
\newblock In \emph{Findings of the Association for Computational Linguistics: ACL-IJCNLP 2021}, pp.\  3621--3634, 2021.

\bibitem[Wang et~al.(2023)Wang, Li, Shao, Xu, Dai, Li, Chen, Y.Wu, and Sui]{Wang2023MathShepherdVA}
Peiyi Wang, Lei Li, Zhihong Shao, Runxin Xu, Damai Dai, Yifei Li, Deli Chen, Y.Wu, and Zhifang Sui.
\newblock Math-shepherd: Verify and reinforce llms step-by-step without human annotations.
\newblock \emph{ArXiv}, abs/2312.08935, 2023.
\newblock URL \url{https://api.semanticscholar.org/CorpusID:266209760}.

\bibitem[Wang et~al.(2024)Wang, Zelikman, Poesia, Pu, Haber, and Goodman]{wang2024hypothesis}
Ruocheng Wang, Eric Zelikman, Gabriel Poesia, Yewen Pu, Nick Haber, and Noah Goodman.
\newblock Hypothesis search: Inductive reasoning with language models.
\newblock In \emph{The Twelfth International Conference on Learning Representations}, 2024.
\newblock URL \url{https://openreview.net/forum?id=G7UtIGQmjm}.

\bibitem[Wang \& Zhou(2024)Wang and Zhou]{wang2024chain}
Xuezhi Wang and Denny Zhou.
\newblock Chain-of-thought reasoning without prompting.
\newblock \emph{arXiv preprint arXiv:2402.10200}, 2024.

\bibitem[Wei et~al.(2022)Wei, Wang, Schuurmans, Bosma, Xia, Chi, Le, Zhou, et~al.]{wei2022chain}
Jason Wei, Xuezhi Wang, Dale Schuurmans, Maarten Bosma, Fei Xia, Ed~Chi, Quoc~V Le, Denny Zhou, et~al.
\newblock Chain-of-thought prompting elicits reasoning in large language models.
\newblock \emph{Advances in neural information processing systems}, 35:\penalty0 24824--24837, 2022.

\bibitem[Xu et~al.(2024)Xu, Fei, Pan, Liu, Lee, and Hsu]{xu2024faithful}
Jundong Xu, Hao Fei, Liangming Pan, Qian Liu, Mong-Li Lee, and Wynne Hsu.
\newblock Faithful logical reasoning via symbolic chain-of-thought.
\newblock \emph{arXiv preprint arXiv:2405.18357}, 2024.

\bibitem[Zhang et~al.(2023)Zhang, Li, Meng, Chang, and Van Den~Broeck]{zhang2023paradox}
Honghua Zhang, Liunian~Harold Li, Tao Meng, Kai-Wei Chang, and Guy Van Den~Broeck.
\newblock On the paradox of learning to reason from data.
\newblock In \emph{Proceedings of the Thirty-Second International Joint Conference on Artificial Intelligence}, pp.\  3365--3373, 2023.

\bibitem[Zhong et~al.(2021)Zhong, Wang, Tang, Xu, Guo, Wang, Yin, Zhou, and Duan]{zhong2021arlsatinvestigatinganalyticalreasoning}
Wanjun Zhong, Siyuan Wang, Duyu Tang, Zenan Xu, Daya Guo, Jiahai Wang, Jian Yin, Ming Zhou, and Nan Duan.
\newblock Ar-lsat: Investigating analytical reasoning of text, 2021.
\newblock URL \url{https://arxiv.org/abs/2104.06598}.

\end{thebibliography}
\bibliographystyle{iclr2025_conference}

\newpage
\appendix


\section{General ability sets Evaluation}
\label{append:General_set}


Here, we present the evaluation results of various models on general datasets. We mainly compare PPO models using our dataset with $\mathrm{PPO}_{\mathrm{UF}}$. For most datasets, there is little difference in performance among the models. On the gsm8k dataset, there is a significant variation in model performance, with some models showing improvement while others experiencing a slight decline.

\begin{table}[h]
\centering
\setlength\tabcolsep{5pt}
\adjustbox{max width=\linewidth}{
\begin{tabular}{@{}lcccccccccc@{}}
\toprule
\textbf{Model} & \textbf{hellaswag} & \textbf{lambada\_openai} & \textbf{mmlu} & \textbf{gsm8k} & \textbf{openbookqa} & \textbf{triviaqa} & \textbf{arc\_easy} & \textbf{arc\_challenge} & \textbf{truthfulqa} & \textbf{average} \\ \midrule

Llama3-8B-Instruct & 57.7 & 71.9 & 63.8 & 32.6 & 34.0 & 51.1 & 81.6 & 52.6 & 43.9 & 54.4 \\ \midrule
$\mathrm{PPO}_{\mathrm{UF}}$ & 57.7 & 71.8 & 64.1 & 37.7 & 34.2 & 44.4 & 81.4 & 53.8 & 47.2 & 54.7 \\
$\mathrm{PPO}_{\mathrm{Na-P}}$ & 57.9 & 71.8 & 64.1 & 34.6 & 34.4 & 47.6 & 81.9 & 54.1 & 46.8 & 54.8 \\
$\mathrm{PPO}_{\mathrm{Fo-P}}$ & 58.0 & 71.7 & 64.1 & 39.0 & 34.2 & 47.3 & 82.2 & 54.4 & 47.3 & 55.4 \\
$\mathrm{PPO}_{\mathrm{Mix-P}}$ & 58.0 & 71.8 & 64.1 & 43.4 & 34.4 & 43.6 & 81.8 & 53.6 & 46.8 & 55.3 \\
$\mathrm{PPO}_{\mathrm{NaO-P}}$ & 57.6 & 71.9 & 64.1 & 33.1 & 34.4 & 52.3 & 81.6 & 52.6 & 42.9 & 54.5 \\
$\mathrm{PPO}_{\mathrm{Na-R}}$ & 57.9 & 71.9 & 64.1 & 41.4 & 34.0 & 45.6 & 81.7 & 54.0 & 47.4 & 55.3 \\
$\mathrm{PPO}_{\mathrm{Na-PN}}$ & 57.8 & 71.8 & 64.1 & 36.6 & 34.4 & 45.4 & 81.6 & 53.9 & 47.0 & 54.7 \\
$\mathrm{PPO}_{\mathrm{Na-P-All}}$ & 58.0 & 71.8 & 64.1 & 40.2 & 34.0 & 44.6 & 81.7 & 53.9 & 47.2 & 55.1 \\
$\mathrm{PPO}_{\mathrm{Na-P-Replace}}$ & 57.9 & 71.8 & 64.0 & 38.2 & 34.0 & 45.7 & 81.6 & 53.7 & 46.8 & 54.9 \\

\bottomrule
\end{tabular}}
    \caption{Performences of PPO Trained Model with different reward model on general ability evaluate sets.}
    \label{tab:all_open_eval_with diffenent_sample_count
    l}
\end{table}

\section{Detail Comparison}
\label{appendix:detail_comparison}

\begin{figure}[H]
\centering
\begin{subfigure}{.32\textwidth}
  \centering
  \includegraphics[width=1.0\linewidth]{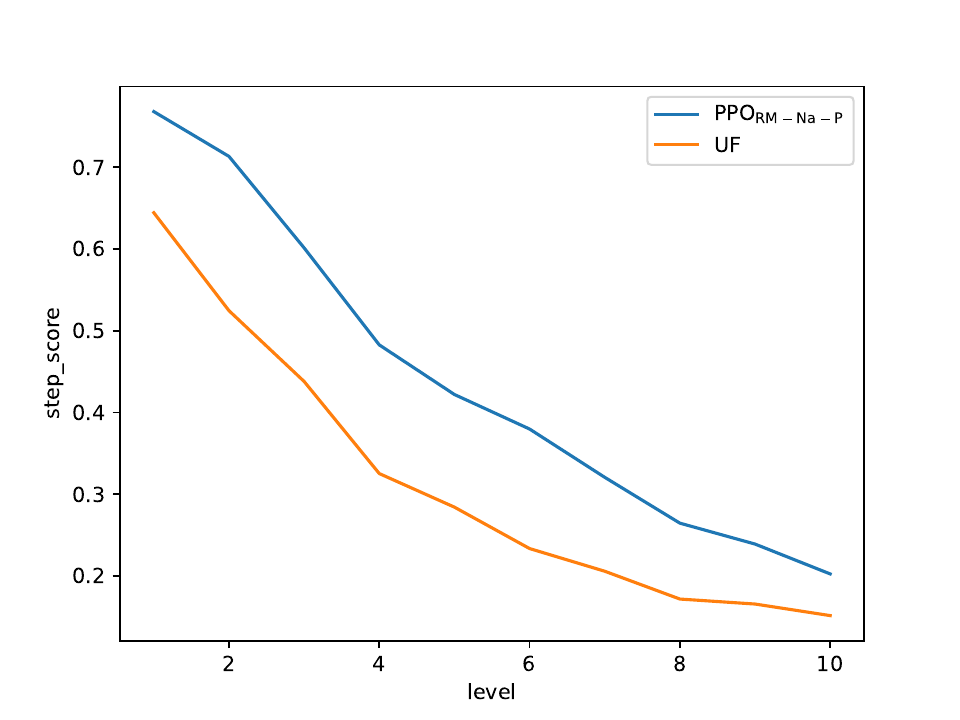}
  \caption{}
  \label{fig:ta_sub1}
\end{subfigure}%
\begin{subfigure}{.32\textwidth}
  \centering
  \includegraphics[width=1.0\linewidth]{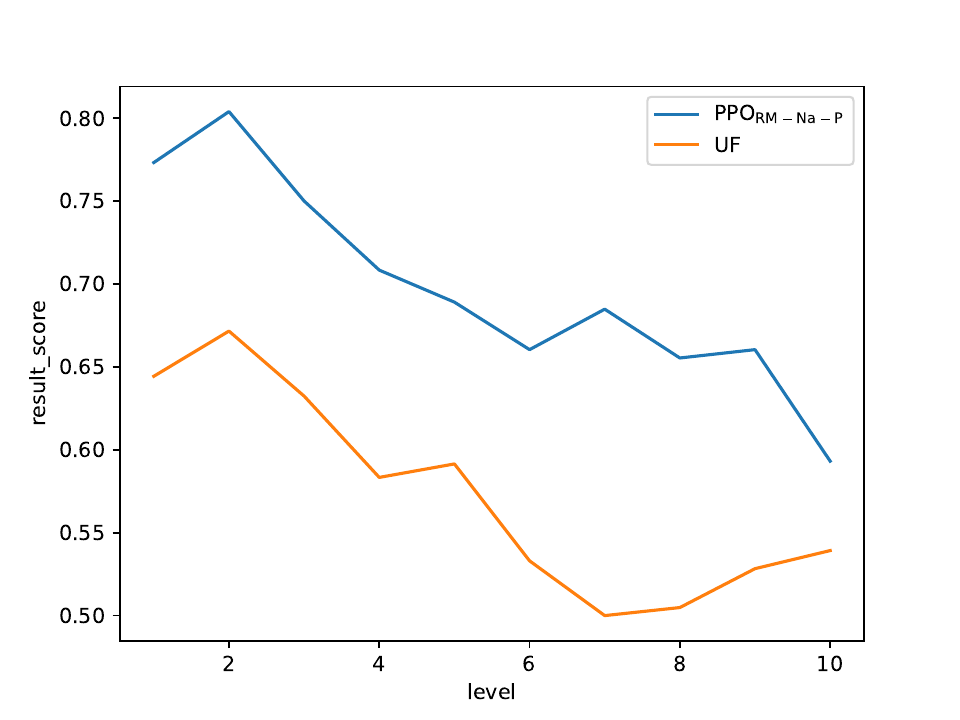}  
  \caption{}
  \label{fig:ta_sub2}
\end{subfigure}
\begin{subfigure}{.32\textwidth}
  \centering
  \includegraphics[width=1.0\linewidth]{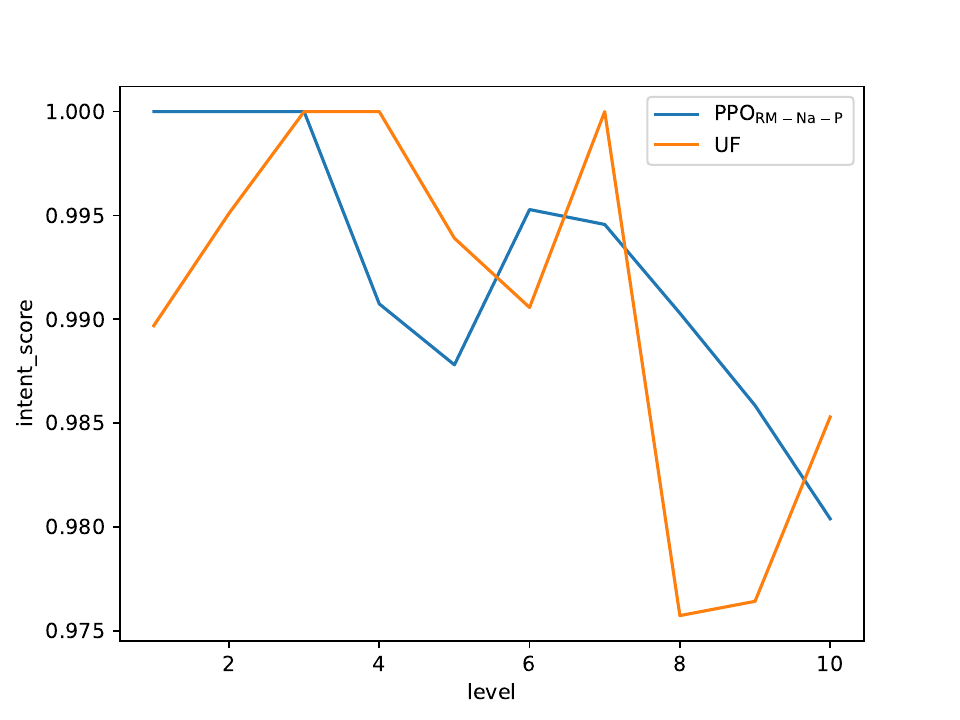}
  \caption{}
  \label{fig:ta_sub3}
\end{subfigure}
\begin{subfigure}{.32\textwidth}
  \centering
  \includegraphics[width=1.0\linewidth]{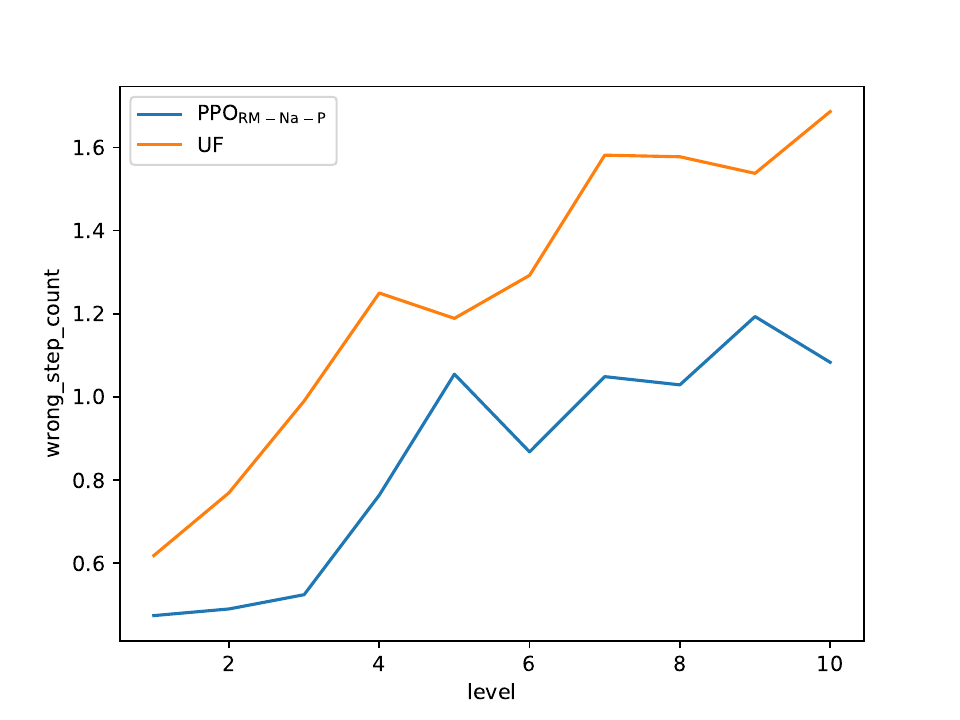}
  \caption{}
  \label{fig:ta_sub4}
\end{subfigure}
\begin{subfigure}{.32\textwidth}
  \centering
  \includegraphics[width=1.0\linewidth]{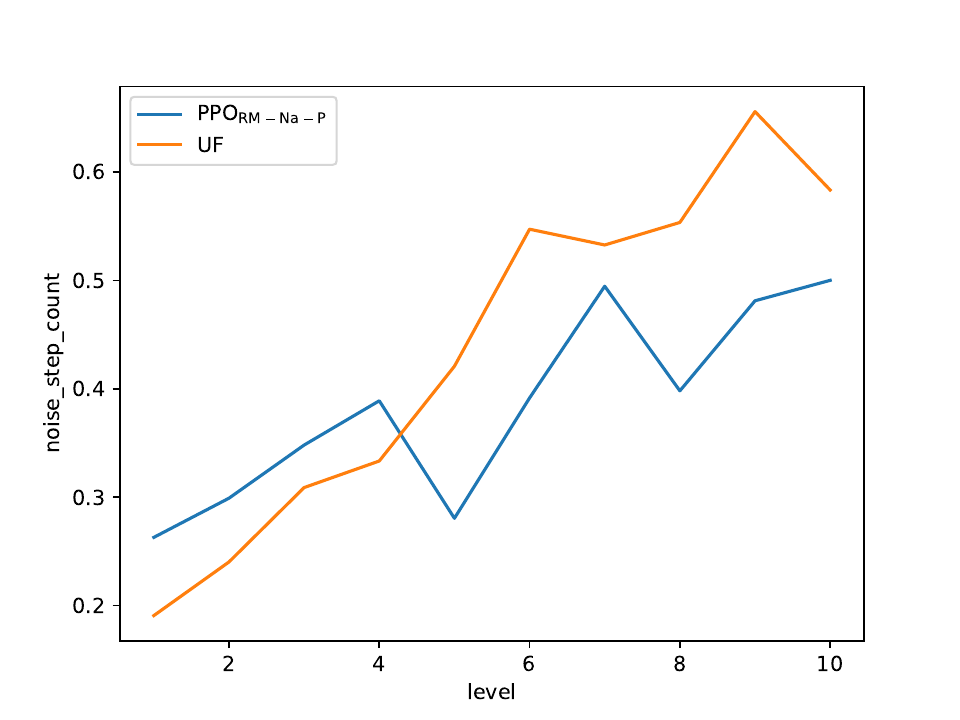}
  \caption{}
  \label{fig:ta_sub5}
\end{subfigure}
\begin{subfigure}{.32\textwidth}
  \centering
  \includegraphics[width=1.0\linewidth]{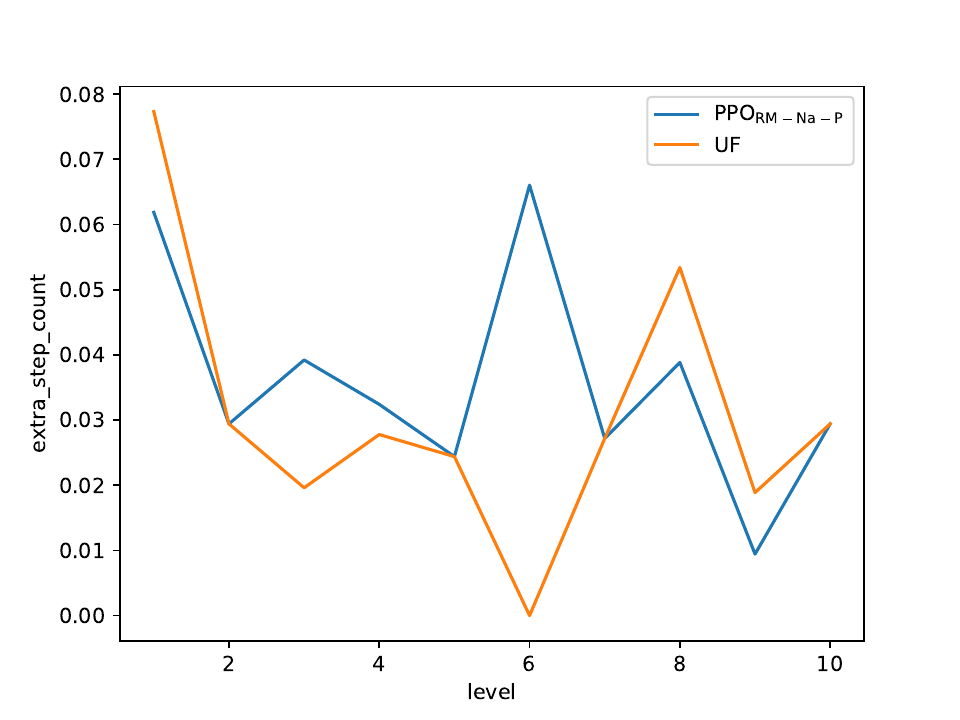}
  \caption{}
  \label{fig:ta_sub6}
\end{subfigure}
\caption{Performances of $\mathrm{PPO}_{\mathrm{UF}}$ and $\mathrm{PPO}_{\mathrm{Na-P}}$ over differnt levels.} 
\label{fig:rm_batch_size_ablation}
\end{figure}

We give some detailed comparison between $\mathrm{PPO}_{\mathrm{UF}}$ and $\mathrm{PPO}_{\mathrm{Na-P}}$. We plot the performances of the two models over different levels, ranging from 1 to 10. Here the level is the necessary steps needed to solve the deductive task. The result is shown in Fig.~\ref{fig:rm_batch_size_ablation}. It can be see that on step and result scores, $\mathrm{PPO}_{\mathrm{Na-P}}$ outperforms $\mathrm{PPO}_{\mathrm{UF}}$ on all levels. $\mathrm{PPO}_{\mathrm{Na-P}}$ also has a lower wrong step count, even through we do not use negative signals when constructing preference data. It can also be see that as the level be larger, performances of models become worse.

\section{Validation Set Accuracy and RewardBench Scores of RMs}
\label{append:rm_acc}

For each training dataset, we split a small set as validation set to test the accuracy of RMs. For logic and UF datasets, we split some prompts so that all RMs haven't been trained on these prompts. We use UF-2667 to denote the validation set of UF dataset with 2667 pairs. For each logic dataset with A responses and B pair method, we denote its validation set with c pairs as ``Logic-A-B-c''. The results are given below.

\begin{table}[h]
\centering
\setlength\tabcolsep{5pt}
\adjustbox{max width=\linewidth}{
\begin{tabular}{@{}lcccccccccc@{}}
\toprule
\textbf{Validation set} & \textbf{RM-UF} & \textbf{RM-Na-P} & \textbf{RM-Fo-P} & \textbf{RM-Mix-P} & \textbf{RM-Na-PN} & \textbf{RM-Na-R} & \textbf{RM-NaO-P} \\ \midrule
UF-2667 & 0.8688 & 0.8635 & \textbf{0.8733} & 0.868 & 0.8695 & 0.8643 & 0.6832 \\
Logic-Na-P-1673 & 0.5415 & 0.7944 & 0.7286 & \textbf{0.7974} & 0.7155 & 0.6629 & 0.7603 \\
Logic-Fo-P-1314 & 0.6058 & 0.777 & \textbf{0.946} & 0.9125 & 0.7154 & 0.6705 & 0.7032 \\ 
Logic-Na-PN-1045 & 0.5898 & 0.7636 & 0.7368 & 0.7531 & \textbf{0.8411} & 0.6852 & 0.6871 \\ 
Logic-Na-R-729 & 0.5617 & 0.8944 & 0.7682 & 0.8971 & 0.882 & \textbf{0.9396} & 0.8203 \\ 
\bottomrule
\end{tabular}}
    \caption{Accuracy of RMs over all validation sets}
    \label{tab:rm_acc}
\end{table}

It can be see that most models has high accuracy on validation sets corresponding to their own training data. RM-UF has low accuracy on our logical data. However, we gain higher accuracy if we add UF dataset to our logical set, as shown by RM-Na-P and RM-NaO-P.

For each dataset, we test these RMs on RewardBench.

\begin{table}[h]
\centering
\setlength\tabcolsep{5pt}
\adjustbox{max width=\linewidth}{
\begin{tabular}{@{}l|cccccccccc@{}}
\toprule
\textbf{RM models} & \textbf{Chat} & \textbf{Chat Hard} & \textbf{Reasoning} & \textbf{Safety} & \textbf{Average Accuracy} \\ \midrule
RM-UF & 0.9665 & 0.5811 & 0.7227 & 0.4972 & 0.6919 \\
RM-Na-P & 0.9749 & 0.6206 & 0.6971 & \textbf{0.5192} & 0.7029 \\
RM-Fo-P & \textbf{0.9832} & 0.6162 & 0.7105 & 0.5176 & 0.7069 \\ 
RM-Mix-P & 0.9693 & \textbf{0.6316} & \textbf{0.7362} & 0.4839 & 0.7052 \\ 
RM-Na-PN & \textbf{0.9832} & 0.6228 & 0.7286 & 0.4888 & 0.7059 \\ 
RM-Na-R & 0.9804 & \textbf{0.6316} & 0.7152 & 0.5046 & \textbf{0.708} \\ 
RM-NaO-P & 0.9134 & 0.4057 & 0.6318 & 0.4083 & 0.5898 \\ 
\bottomrule
\end{tabular}}
    \caption{Accuracy of RMs over all validation sets}
    \label{tab:rm_rewardbench}
\end{table}

We can check the reasoning score of RewardBench on all the RMs with the final performances of PPOs. We find that this reasoning scores can not be a proper standard for RM chosen. This might be caused by the mismatch of both prompts and responses.

\section{Evaluation Sets Details}

We list all evaluation sets we used below.
\begin{itemize}[]
    \item \textbf{lambada\_openai}: Tasks designed to predict the endings of text passages, testing language prediction skills.\cite{lambada_openai}
    \item \textbf{mmlu}: Massive Multitask Language Understanding benchmark for broad domain language evaluation. Several variants are supported.\cite{Hendrycks_Burns_Basart_Zou_Mazeika_Song_Steinhardt_2020}
    \item \textbf{gsm8k}: A benchmark of grade school math problems aimed at evaluating reasoning capabilities.\cite{cobbe2021training}
    \item \textbf{openbookqa}: Open-book question answering tasks that require external knowledge and reasoning.\cite{OpenBookQA2018}
    \item \textbf{triviaqa}: A large-scale dataset for trivia question answering to test general knowledge.\cite{JoshiTriviaQA2017}
    \item \textbf{arc\_easy}: Tasks involving complex reasoning over a diverse set of questions.\cite{Clark2018ThinkYH}
    \item \textbf{arc\_challenge}: Tasks involving complex reasoning over a diverse set of questions.\cite{Clark2018ThinkYH}
    \item \textbf{truthfulqa}: A QA task aimed at evaluating the truthfulness and factual accuracy of model responses.\cite{lin-etal-2022-truthfulqa}
    \item \textbf{logicqa}: Logical reasoning tasks requiring advanced inference and deduction.\cite{liu2020logiqa}, 
    \label{sec:common_evaluate}
\end{itemize}

\section{Evaluate Set Template}
\subsection{AR-LSAT}
\label{app:ar_lsat_demo}

\begin{tcolorbox}  
    "id": "ar\_lsat\_201609\_3-G\_4\_23",
    
    "context": "There are exactly six computers—P, Q, R, S, T, and U—on a small network. Exactly one of those computers was infected by a virus from outside the network, and that virus was then transmitted between computers on the network. Each computer received the virus exactly once. The following pieces of information concerning the spread of the virus have been established: No computer transmitted the virus to more than two other computers on the network. S transmitted the virus to exactly one other computer on the network. The computer that transmitted the virus to R also transmitted it to S. Either R or T transmitted the virus to Q. Either T or U transmitted the virus to P.",
    "question": "If P is the only computer that transmitted the virus to two other computers on the network, which one of the following must be true?",
    
    "options": [
    
      "A) S transmitted the virus to T.",
      
      "B) T transmitted the virus to P.",
      
      "C) Q did not transmit the virus to any other computer on the network.",
      
      "D) R did not transmit the virus to any other computer on the network.",
      
      "E) U did not transmit the virus to any other computer on the network."
      
    ],
    
    "answer": "C"
\end{tcolorbox}

\section{Templates for proof and judgement prompts}
\label{app:prompt_template}

We use below templates to construct our prompts.

The proof template is
\begin{tcolorbox}
Given:\\
\{Inference conditions.\}\\
Prove: \{conclusion\}.
\end{tcolorbox}

The judgement template is

\begin{tcolorbox}
We have:\\
\{Inference conditions.\}\\
Show \{conclusion\} is correct or not.
\end{tcolorbox}

\section{Templates for formatted response generation}
\label{app:format_template}

To generate response with JSON format, we use few-shot learning to instruct model to generate. The template is given below. The template for proof problem is.

The template for judgement problem is given below. The template for proof problem is quite similar.
\begin{tcolorbox}
I will give you a few given conditions and you need to check whether a given conclusion is correct or not based on these conditions.

You need to list all the deductive process in a json style. For each step, you need to list:\\
* condition: the conditions you use to conduct deduction,\\
* conclusion: the conclusion you get,\\
* format conclusion: a dictionary which has below three terms:\\
    ** Subject: the subject of your conclusion, which should be an affirmed noun.\\
    \\
    ** Predication: the prediate of your conclusion, which should be an affirmed noun.\\
    ** type: which is one in ['A','E','I','O']. The type of one proposition with subject $S$ and predicate $P$:\\
        *** Type 'A': 'all $S$ are $P$', or '$S$ is $P$'.\\
        *** Type 'E': 'None of $S$ is $P$', or '$S$ is not $P$'.\\
        *** Type 'I': 'There exists one $S$ that is $P$'.\\
        *** Type 'O': 'There exists one $S$ that is not $P$'.\\
Finally you should give a 'result' if you are required to check whether the given conclusion is correct or not. If it is correct, return 'Correct'; otherwise, return 'Wrong'.\\
\\
Your answer should be return with below format:\\
\{`steps': [\\
        `condition': [`xxx', `xxx'],\\
        `conclusion': [`xxx'],\\
        `format\_conclusion': \{`Subject': `xxx', `Predicate': `xxx', `type', `x'\}\\
    ],[\\
        `condition': [`xxx', `xxx'],\\
        `conclusion': [`xxx'],\\
        `format\_conclusion': \{`Subject': `xxx', `Predicate': `xxx', `type', `x'\}\\
    ],[\\
        `condition': [`xxx', `xxx'],\\
        `conclusion': [`xxx'],\\
        `format\_conclusion': \{`Subject': `xxx', `Predicate': `xxx', `type', `x'\}
    ],\\
    `result': `xxx'\\
\}\\
\\
Examples:\\
\{Examples\}\\
\\
\#\#Input:\\
\{Prompt\}\\
\#\#Output:
\end{tcolorbox}

\section{Detailed Evaluation Results}
\label{app:detailed_eval}

Here we give detailed evaluation results form LLMs on our MuseD dataset. We show the step scores, which we think is the most import score, over the deductive steps and the noise conditions.

\begin{table}[H]
\centering
\setlength\tabcolsep{5pt}
\adjustbox{max width=\linewidth}{
\begin{tabular}{@{}l|c|ccccccccc@{}}
\toprule
\textbf{Levels} & \textbf{count} & \textbf{GPT-4} & \textbf{GPT-4o} & \textbf{GPT-4o-mini} & \textbf{GPT-4-o1-preview} & \textbf{GPT-4-o1-mini} & \textbf{Qwen2-72B-Instruct} & \textbf{Qwen2.5-72B-Instruct} & \textbf{Llama3.1-72B-Instruct}\\ \midrule
1 & 194 & 0.88 & \textbf{0.93} & 0.74 & \textbf{0.93} & 0.83 & 0.89 & 0.92 & 0.91\\
2 & 204 & 0.89 & 0.94 & 0.66 & \textbf{0.96} & 0.91 & 0.89 & 0.95 & 0.88\\
3 & 204 & 0.83 & 0.92 & 0.59 & \textbf{0.94} & 0.88 & 0.84 & \textbf{0.94} & 0.82\\
4 & 216 & 0.75 & 0.86 & 0.49 & \textbf{0.9} & 0.82 & 0.78 & 0.89 & 0.78\\
5 & 164 & 0.79 & 0.86 & 0.46 & \textbf{0.87} & 0.82 & 0.73 & \textbf{0.87} & 0.84\\
6 & 212 & 0.71 & 0.8 & 0.41 & \textbf{0.87} & 0.76 & 0.73 & 0.84 & 0.74\\
7 & 184 & 0.64 & 0.73 & 0.32 & 0.77 & 0.68 & 0.66 & \textbf{0.78} & 0.7\\
8 & 206 & 0.65 & 0.72 & 0.31 & \textbf{0.81} & 0.68 & 0.64 & 0.73 & 0.65\\
9 & 212 & 0.61 & 0.69 & 0.27 & \textbf{0.76} & 0.66 & 0.6 & 0.7 & 0.66\\
10 & 204 & 0.56 & 0.64 & 0.26 & \textbf{0.7} & 0.58 & 0.55 & 0.63 & 0.58\\
\bottomrule
\end{tabular}}
    \caption{Performances of Different Models on MuseD over prompt levels.}
    \label{tab:mused_eval_levels}
\end{table}
\begin{table}[H]
\centering
\setlength\tabcolsep{5pt}
\adjustbox{max width=\linewidth}{
\begin{tabular}{@{}l|c|ccccccccc@{}}
\toprule
\textbf{Levels} & \textbf{count} & \textbf{GPT-4} & \textbf{GPT-4o} & \textbf{GPT-4o-mini} & \textbf{GPT-4-o1-preview} & \textbf{GPT-4-o1-mini} & \textbf{Qwen2-72B-Instruct} & \textbf{Qwen2.5-72B-Instruct} & \textbf{Llama3.1-72B-Instruct}\\ \midrule
0 & 218 & 0.82  & \textbf{0.93} & 0.65 & 0.91 & 0.83 & 0.86 & 0.91 & 0.86\\
1 & 484 & 0.82  & \textbf{0.91} & 0.58 & 0.9 & 0.84 & 0.83 & \textbf{0.91} & 0.83\\
2 & 546 & 0.73  & 0.82 & 0.44 & \textbf{0.87} & 0.77 & 0.76 & 0.84 & 0.76\\
3 & 394 & 0.69  & 0.74 & 0.35 & \textbf{0.82} & 0.72 & 0.67 & 0.77 & 0.71\\
4 & 228 & 0.62  & 0.68 & 0.3 & \textbf{0.77} & 0.70 & 0.58 & 0.72 & 0.65\\
5 & 100 & 0.53  & 0.62 & 0.25 & \textbf{0.74} & 0.57 & 0.52 & 0.66 & 0.6\\
6 & 26 & 0.65  & 0.68 & 0.23 & \textbf{0.78} & 0.68 & 0.42 & 0.53 & 0.54\\
7 & 4 & 0.42  & 0.51 & 0.14 & 0.53 & 0.56 & 0.32 & \textbf{0.58} & 0.41\\
\bottomrule
\end{tabular}}
    \caption{Performances of Different Models on MuseD over noise counts.}
    \label{tab:mused_eval_noise}
\end{table}

As the result shows, LLMs become worse as the the level becomes larger. That is, the more complex the prompt is, the lower the step score is. GPT-4-o1-preview still keeps a 0.7 step score for level 10. Similarly, the more noise conditions are involved, the worse performances LLMs have.

\end{document}